\documentclass[11pt]{article}

\usepackage[final]{acl}

\usepackage{times}
\usepackage{latexsym}

\usepackage[T1]{fontenc}
\usepackage[utf8]{inputenc}

\usepackage{microtype}

\usepackage{inconsolata}

\usepackage{graphicx}
\usepackage{multirow}
\usepackage{booktabs}
\usepackage{array}
\usepackage{arydshln}
\usepackage{fvextra}
\usepackage{placeins}
\usepackage{longtable}
\graphicspath{{../counterfactual-faithfulness/analysis/judge_comparison/}{../counterfactual-faithfulness/analysis/judge_comparison/fluency/}{./}}

\newcommand\Tstrut{\rule{0pt}{2.3ex}}       % "top" strut
\newcommand\Bstrut{\rule[-0.9ex]{0pt}{0pt}} % "bottom" strut

\title{Do LLMs Make More Mistakes If They Do Not Believe the Input Data?}

\author{
Peter Kochelka, 
Aleš Manuel Papáček,
Vojtěch Dvořák \and
Ondřej Dušek   \\
Charles University, Faculty of Mathematics and Physics \\
Institute of Formal and Applied Linguistics \\
Prague, Czech Republic \\
\texttt{\{kochelka,papacek,dvorak,odusek\}@ufal.mff.cuni.cz}
}

\begin{document}
\maketitle
\begin{abstract}
Large language models (LLMs) are prone to hallucinating or misinterpreting facts, which impairs their usability in retrieval-augmented generation or data-to-text systems. We analyse how faithfulness of LLMs to provided context depends on how plausible they perceive the context to be (context--memory conflict). To better identify error patterns, we make use of the increased difficulty of non-English and low-resource language text generation and input data based on local knowledge, only partially captured in models' parametric knowledge. We let the models generate text in English, Czech, Slovak and Upper Sorbian from factual (FA), counterfactual (CFA) and fictional (FI) RDF triples containing local Czech and Slovak data.
%We evaluate the generations with two LLM judges, Kimi K3 and DeepSeek V4 Pro Preview, and validate their faithfulness scores on a stratified human-annotated sample.
Contrary to our expectations, we observe only a weak context--memory conflict on the human-annotated sample. For Kimi K3 as an LLM judge, which agrees well with human annotations on the sample, counterfactual inputs receive only slightly lower faithfulness scores than factual ones ($-0.05$ on a 1--5 scale).
%DeepSeek V4 Pro Preview reports the same effect three to four times larger, 
We also find that a suboptimal choice of LLM judge would lead to overestimating the strength of the context--memory conflict.
\end{abstract}

\section{Introduction}
\label{sec:introduction}

LLMs can generate fluent text from structured inputs with little task-specific training, as shown by \citet{axelsson-skantze-2023-using} for GPT-3.5 and improving with newer, larger models. However, fluent output can still omit, alter or add facts from the provided data \cite{kasnerTraditionalBenchmarksAnalyzing2024}.

\citet{xuKnowledgeConflictsLLMs2024} describe the \emph{context--memory conflict}, where a model's parametric knowledge conflicts with the provided context.
A model may correct or ignore information it perceives as erroneous.
To analyse this tension, we focus on counterfactual and fictional RDF\footnote{Triples in the form '\texttt{subject | predicate | object}'} inputs.

Existing benchmarks may not fully reveal this tension due to training data leakage \citep{songMultilingualVerbalisationKnowledge2025}.
We therefore create factual (FA), counterfactual (CFA) and fictional (FI) groups of RDF triples from local (Czech and Slovak) knowledge data \citep{libovickyCUSQALocalKnowledgeOrientedOpenEnded2026}, as local knowledge is less likely to be captured by models' parametric knowledge. %of current open-weight models. 
We ask LLMs to classify each input as FA, FI or CFA. This way, we obtain input plausibility %of every input
as perceived by the model, which allows us to compare its behaviour based on the context-memory conflict strength. We then generate sentences from these triple groups in English, Czech, Slovak and Upper Sorbian using open-weight LLMs of various sizes, and we analyse differences based on perceived (and actual) input factuality. 

We find that larger models classify input factuality more accurately.
With Kimi K3 as the primary LLM judge based on its agreement with human annotations, we observe only a mild context--memory conflict (rating counterfactuals items only $0.05$ points lower on a 1--5 scale compared to factual).
However, DeepSeek V4 Pro Preview reports the same effect three to four times larger, so a wrong choice of LLM judge can lead to overestimating the strength of the context--memory conflict.
%The broad model-size and language trends are stable across two LLM judges, but the counterfactual effect itself is not: on matched items it is $-0.05$ on a 1--5 scale under the judge that agrees better with our human annotations, against $-0.18$ under the other. Human validation shows that the judges are themselves mildly more lenient towards factual inputs, by several times that gap, and that counterfactual and fictional triples cost about as much fluency as faithfulness, even though the fluency judge only sees set of subjects and objects, not the triples themselves.
We release our code and data including annotations at \url{https://github.com/pkochelka/counterfactual-faithfulness}.

\section{Related Work}

\citet{axelsson-skantze-2023-using} showed that prompted LLMs can verbalize knowledge graphs;
%without task-specific training. 
this also expands to under-resourced languages %by 
\cite{lorandiHighqualityDatatoTextGeneration2024}.
The GEM 2024 shared task \citep{GEM2024} evaluated factual, counterfactual, and fictional inputs. 
\citet{sunTaskMattersKnowledge2026}
show that model behaviour with respect to the context--memory conflict \citep{xuKnowledgeConflictsLLMs2024}
depends on the task framing,
and that models may still rely on parametric knowledge even when instructed to
use the context. Studying this is difficult with established
benchmarks, whose data may appear in pretraining \citep{songMultilingualVerbalisationKnowledge2025}. 
To counteract this, \citet{kasnerTraditionalBenchmarksAnalyzing2024} evaluate data-to-text
generation using data collected on-the-fly.
\citet{libovickyCUSQALocalKnowledgeOrientedOpenEnded2026} and \citet{MULTILOKO}
show a similar effect on question-answering by using relatively obscure local knowledge of non-English-speaking nations.
For evaluation, LLM judges can offer better correlation than older automatic metrics \cite{xu-etal-2023-instructscore} but this  varies across tasks \cite{bavaresco-etal-2025-llms} and is prone to biases \cite{zhengJudgingLLMasaJudgeMTBench2023}.

Our work combines the above perspectives of multilingual and low-resource data-to-text, context-memory conflicts and less familiar data by measuring whether a model’s own perception of factuality affects its faithfulness when verbalizing factual, counterfactual, and fictional RDF inputs.
We use LLM judges audited on human-annotated samples.

% TODO: add a short LLM-as-judge paragraph here -- our headline result is that
% the effect size depends on the judge, so judge reliability deserves related
% work. E.g. \citet{zhengJudgingLLMasaJudgeMTBench2023}

% Navrh:
% Using LLM-as-judge offer a scalable alternative to human evaluation but remain susceptible to biases such as self-enhancement \citep{zhengJudgingLLMasaJudgeMTBench2023}. This motivates validating judge selection against human annotations, especially when it affects the conclusions.

\section{Data}
\label{sec:data}

We use the Czech and Slovak portions of the local-knowledge CUS-QA benchmark \citep{libovickyCUSQALocalKnowledgeOrientedOpenEnded2026}. 
It is built from native-speaker questions grounded in local Wikipedia content.
We convert it into RDF triples and add counterfactual and fictional versions (see Appendix Figure~\ref{fig:cusqa}).

\paragraph{Converting CUS-QA into triples}

We feed question-answer pairs to Mistral Medium 3.5~\citep{mistralMistralMedium35ModelCard2026}, prompting it to generate a statement and corresponding triples for each pair. Each subject and object in the resulting triples is assigned a type by gpt-oss-120b \citep{openaiGptOssModelCard2025} using a constrained label set (e.g., personal name, place, date, number). Gpt-oss-120b is given context from Wikipedia to improve type accuracy.
The result is our factual (FA) dataset.

\paragraph{Counterfactual and fictional sets}
are created following \citet{GEM2024}. Counterfactual (CFA) data are built from factual triples by randomly substituting subjects and objects with other entities of the same type. 
Fictional (FI) data are created similarly, using fictional entities generated by Claude Sonnet 5 \citep{anthropic2026claudesonnet5} and gpt-oss-120b.

\medskip\noindent
The resulting dataset contains 2,847 instances, each composed of 1-9 RDF triples: 1,482 with Czech local data (530 FA, 476 CFA, and 476 FI) and 1,365 with Slovak (493 FA, 436 CFA, and 436 FI).

\section{Experiments}

\paragraph{Classification.}

To test models' parametric knowledge, we prompt them to classify the input set of triples into FA/CFA/FI (cf.~Section~\ref{sec:data}).
The classification is repeated five times with a temperature
of 1.0, the majority label is used as the final prediction. We also measure model consistency (proportion of unanimous classifications).

\paragraph{Data-to-text generation.}

We prompt the LLMs to verbalize each RDF triple set as one or a few natural sentences in the target language, using greedy decoding.

\medskip\noindent
Full prompts for both tasks are in Appendix~\ref{sec:appendix-prompts}.

\subsection{Experimental Settings}
\label{sec:exp-settings}

We use Czech and Slovak as the data languages and Czech, Slovak, English, and
Upper Sorbian as the prompt/target languages. Czech and Slovak let us compare prompts that match the language/locale of the input data (“Same local”) with prompts in the other closely related language (“Other local”). English represents a very high-resource language,
whereas Upper Sorbian represents a language closely related to Czech and Slovak but very low-resource.\footnote{Upper Sorbian is estimated to have around 25,000 speakers \citep{minorityrights_sorbs}.}
The experiment thus runs along four dimensions: model (nine models described in Section~\ref{sec:models}), data language (\textsc{CS}, \textsc{SK}), example variant (\textsc{CFA}, \textsc{FA}, \textsc{FI}; Section~\ref{sec:data}), and prompt/target language (\textsc{CS}, \textsc{SK}, \textsc{EN}, \textsc{HSB}), yielding 11,388 generations per model.

\subsection{Tested Models}
\label{sec:models}

We evaluate nine open-weight generator models from five model families, separated into three groups:
\emph{Large} (>100B) -- Qwen3.5 122B-A10B \cite{qwenQwen35ModelCard2026} and gpt-oss-120b~\cite{openaiGptOssModelCard2025}; 
\emph{Medium} (>4B) -- Gemma 4 31B and E4B \cite{gemmaTeamGemma4Technical2026}, 
Llama 4 Scout 17B-16E Instruct \cite{metaLlama4ScoutModelCard2025}
and Qwen3.5 9B;  
\emph{Small} -- Gemma 4 E2B, Tiny Aya Global 3.35B \cite{salamancaTinyAya2026} and Qwen3 1.7B \cite{yangQwen3TechnicalReport2025}.
Configuration details are in Appendix~\ref{sec:model-config}.

\subsection{LLM-as-Judge Evaluation}

As the full dataset is too large to score manually, we use LLM judges to rate
each generated text from 1 to 5 for \textbf{faithfulness} to the input RDF
triples and \textbf{fluency} in the target language. 
%We initially used DeepSeek V4 Pro Preview~\citep{deepseekDeepseekV4ProModelCard2026}, but human validation showed weaker agreement than several alternatives. We therefore use Kimi K3~\citep{moonshotKimiK3ModelCard2026}, which offers the best practical balance of agreement and cost, as our primary judge. We retain DeepSeek for comparison to demonstrate how judge-specific biases can affect the resulting scores and conclusions.

\paragraph{Faithfulness} measures agreement with the source RDF triples, penalizing omissions and unsupported claims.
We also test how closely the model followed the instructions to provide coherent text in the target language, penalizing outputs in a different language (e.g. Czech instead of Slovak) or non-natural language (e.g., leaving triples as-is).
The judge also identifies the triple(s) that were not verbalized correctly and provides a short comment.

\paragraph{Fluency} measures only how natural-sounding the sentences are in the target language, regardless of how faithful they are. The judge is not provided the input triples in order to focus on fluency only; however, flattened entities from the source triples are provided to adjust expectations on proper names. 

\medskip\noindent
Full prompts are in Appendix~\ref{sec:appendix-prompts}. We report mean scores and bootstrap 95\% confidence intervals.%
\footnote{For population-level mean scores, brackets report 95\% confidence intervals
from 10{,}000 source-item cluster-bootstrap resamples. Generated sentences from the same source triples are kept together during resampling. Paired contrasts instead resample matched dataset--item
clusters. \label{fn:bootstrap}}

% \paragraph{Worked example.} 
% Given English as the target language and "\texttt{Vyškov | locatedIn | Turnov}" as the triple from the Czech CFA dataset, one model generated "\textit{Vyškov is located in Turnov}", which is a faithful and fluent sentence. Another model generated "\textit{Turnov is located in Vyškov}", which is fluent but reverses the relation and is therefore unfaithful.

\subsection{Manual Annotation for Judge Tuning}\label{sec:judge-tuning}

To obtain reference labels against which we can tune and audit the LLM judge prompt, we pick a small tuning sample (540 instances) for manual annotation by the authors.
We balance it by factuality, model and language.
We picked DeepSeek V4 Pro Preview \cite{deepseekDeepseekV4ProModelCard2026} and ran it with an initial prompt to also balance the sample with respect to faithfulness scores (see Appendix~\ref{sec:manual-annot-details}). 
We hid the initial judge scores and LLM identifiers for annotation.
We also annotated a heldout sample of 108 instances for additional LLM judge evaluation, using the same stratified design.
We then iteratively tuned the judge prompt with DeepSeek on the tuning sample to more make it closely follow the guidelines and match the human labels. 
%Every judge score reported in this paper comes from the frozen final prompt.

\subsection{Judge Validation}

Using the final optimized prompt on both the tuning and heldout samples, we evaluated DeepSeek as well as four alternatives: Kimi K3 \cite{moonshotKimiK3ModelCard2026}, Claude Opus 5 \cite{anthropic2026claudeopus5} with a low reasoning effort and GPT-5.6 Sol \cite{openai2026gpt56sol} with low and high reasoning.
%We tuned the judge prompt with DeepSeek on 540 author-annotated outputs. After freezing the prompt, we compared five judges on this sample and additionally on a disjoint sample of 108 outputs drawn using the same stratified design. 

\begin{table}[t]
\centering
\small\setlength{\tabcolsep}{2pt}
\begin{tabular}{lcrrr>{\hspace{3mm}}cr}
\toprule
& \multicolumn{4}{c}{Tuning sample (540)}
& \multicolumn{2}{c}{Heldout (108)} \\
\cmidrule(r){2-5}\cmidrule(l){6-7}
Judge & $\kappa_w$ & Exact & $\leq 1$ & MAE
      & $\kappa_w$ & Exact \\
\midrule
Claude\,Opus\,5\,(lo)\hspace{-0.5mm}
  & \textbf{0.776} & \textbf{67.6} & \textbf{89.4} & \textbf{0.483}
  & \textbf{0.755} & 70.4 \\
GPT-5.6 Sol (hi)
  & 0.744 & 66.7 & 88.7 & 0.517
  & 0.739 & 69.4 \\
Kimi K3
  & 0.712 & 65.7 & 88.3 & 0.541
  & 0.711 & \textbf{71.3} \\
GPT-5.6 Sol (lo)
  & 0.700 & 65.4 & 87.6 & 0.567
  & 0.750 & \textbf{71.3} \\
DeepSeek
  & 0.555 & 54.8 & 80.9 & 0.793
  & 0.350 & 48.1 \\
\bottomrule
\end{tabular}
\caption{Agreement with human faithfulness labels. $\kappa_w$ is
quadratic-weighted kappa, Exact and $\leq 1$ are percentages, and MAE is mean
absolute error. %All five judges cover the same 540-row tuning sample and the
%same disjoint 108-row held-out sample. Opus 5 is Claude Opus 5
%\citep{anthropic2026claudeopus5} at low reasoning effort; Sol is GPT-5.6 Sol
%\citep{openai2026gpt56sol}, with reasoning effort in parentheses.
Parentheses show the reasoning effort setting.
}
\label{tab:human-judge-agreement}
\end{table}

Table \ref{tab:human-judge-agreement} shows agreement with human annotation of faithfulness.
Claude Opus 5 agrees best on both samples, but
is prohibitively expensive at full scale. Kimi offers the best balance of
agreement and cost. Its agreement remains stable on the heldout sample
($\kappa_w$ 0.712 $\rightarrow$ 0.711), whereas DeepSeek degrades substantially
(0.555 $\rightarrow$ 0.350), suggesting overfitting and weak generalizability. 
We thus pick Kimi as our primary judge.

We further compared Kimi and DeepSeek on fluency annotation, with Kimi agreeing better than DeepSeek ($\kappa_w$ 0.553 versus 0.456).
%\paragraph{Human validation of the fluency judges.}
%We also annotated fluency on the same samples, but restrict this validation to
%English, Czech and Slovak: none of the authors speaks Upper Sorbian well enough
%to rate naturalness reliably, so the partial Upper Sorbian fluency labels
%collected early in annotation are excluded from it. On the 475
%English, Czech and Slovak rows with a human fluency label, Kimi again agrees
%better than DeepSeek ($\kappa_w$ 0.553 versus 0.456; exact 40.4\% versus
%32.0\%), and the held-out sample separates them further ($\kappa_w$ 0.756
%versus 0.470). 
Both fluency judges are stricter than the annotators
(bias $-0.50$ for Kimi and $-0.72$ for DeepSeek), and this is
language-dependent. % reaching $-0.76$ and $-0.88$ on Slovak. 
Absolute fluency
levels and small cross-language fluency differences should therefore be read
with caution. % than the faithfulness scores.
Appendix~\ref{sec:judge-audit-details}
provides additional details on judge validation. 
%DeepSeek robustness tables, fluency judge agreement, reasoning-effort comparisons, and population-scale disagreement analysis.

\section{Results}
\label{sec:results}

\subsection{Classification}

Table~\ref{tab:model-results} shows that most models classify the majority of inputs as FA or CFA; Tiny Aya is the exception, classifying most as FI.
Small models show limited knowledge of the local facts, with 33--36\% accuracy, close to the ${\sim}33\%$ random baseline. As the model size grows, so does accuracy.

\begin{table}[t]
\centering
\def\pz{\phantom{0}}
\small\setlength{\tabcolsep}{4pt}
\begin{tabular}{@{}lccc>{\hspace{1mm}}cc@{}}
\toprule
Model & Acc. & Unan. & FA/FI/CFA & Faith. & Flu. \\
\midrule
Qwen3.5 122B   & 52.0 & 81.6 & 40/\pz9/51 & 4.90 & \textbf{4.15} \\
GPT OSS 120B & \textbf{54.2} & 64.4 & 36/16/48 & \textbf{4.96} & 3.98\Bstrut \\
\hdashline[0.5pt/2pt]
Gemma4 31B   & 52.8 & \textbf{98.5} & 55/\pz6/38 & 4.94 & 3.76\Tstrut \\
Llama4 17B    & 48.4 & 71.2 & 56/11/33 & 4.77 & 4.09 \\
Qwen3.5 9B  & 47.3 & 28.4 & 53/19/28 & 4.84 & 4.13 \\
Gemma4 E4B   & 37.5 & 89.9 & 92/\pz4/\pz4 & 4.46 & 3.49\Bstrut \\
\hdashline[0.5pt/2pt]
Gemma4 E2B   & 36.2 & 93.4 & 94/\pz1/\pz5 & 4.12 & 2.84\Tstrut \\
Tiny Aya 3.35B   & 33.1 & 19.3 & \pz9/78/13 & 3.66 & 3.74 \\
Qwen3 1.7B  & 35.3 & 53.5 & 74/20/\pz6 & 3.49 & 3.00 \\
\bottomrule
\end{tabular}
\caption{Classification and generation results. Acc.\ = exact majority-vote
classification accuracy, Unan.\ = \% of examples with unanimous classification in five runs. FA/FI/CFA is the percentage distribution of
 majority predictions across classes (excl.\ 59 unparseable
 classification attempts). Faith.\ and Flu.\ = mean Kimi K3 scores on a 1--5
 scale over all 11,388 outputs per model. Model size groups are split by dashed lines.}
\label{tab:model-results}
\end{table}

\subsection{Generation}
\label{sec:generation-results}

Table~\ref{tab:model-results} reports the results of each tested model including faithfulness and fluency as given by the Kimi K3 judge. Faithfulness generally increases with model size, while fluency follows a less consistent pattern. 
%DeepSeek gives a lower overall mean than Kimi (4.32 versus 4.46) but preserves the broad model-size trend.
Faithfulness also generally decreases
with the number of input triples under both judges (see Appendix
Table~\ref{tab:faithfulness-triples}).

%We group prompts into English (EN), Upper Sorbian (HSB), Same local language
%(Czech prompts with Czech data and Slovak prompts with Slovak data), and Other
%local language (the opposite pairings); EN and HSB pool both source datasets.

Table~\ref{tab:language-condition-results} presents Kimi faithfulness by
prompt-language condition (cf.~Section~\ref{sec:exp-settings}). English and matching local-language prompts perform
best, non-matching Czech and Slovak prompts perform slightly worse, and Upper
Sorbian is the most difficult, especially for smaller models.
Upper Sorbian accounts for
50.4\% of faithful-but-disfluent cases, and large models retain mean
faithfulness 4.85 in Upper Sorbian while fluency falls to 3.53, a 1.32-point
gap compared with 0.54 in English. Similarly, changing from matching to the other local language drops faithfulness by 0.03 but fluency by 0.29.

\begin{table}[t]
\centering
% \scriptsize
% \begin{tabular}{lcccc}
% \toprule
% Models & EN & Same & Other & HSB \\
% \midrule
% Large & \textbf{4.97} & \textbf{4.97} & \textbf{4.92} & \textbf{4.85} \\
%  & {\scriptsize [4.97, 4.98]} & {\scriptsize [4.96, 4.97]} & {\scriptsize [4.90, 4.93]} & {\scriptsize [4.83, 4.87]} \\
% Medium & 4.94 & 4.91 & 4.87 & 4.29 \\
%  & {\scriptsize [4.94, 4.95]} & {\scriptsize [4.90, 4.92]} & {\scriptsize [4.86, 4.88]} & {\scriptsize [4.26, 4.31]} \\
% Small & 4.70 & 4.45 & 4.44 & 1.43 \\
%  & {\scriptsize [4.69, 4.72]} & {\scriptsize [4.43, 4.47]} & {\scriptsize [4.41, 4.46]} & {\scriptsize [1.41, 1.45]} \\
% \midrule
% Overall & 4.87 & 4.77 & 4.74 & 3.46 \\
%  & {\scriptsize [4.86, 4.88]} & {\scriptsize [4.76, 4.78]} & {\scriptsize [4.72, 4.75]} & {\scriptsize [3.44, 3.47]} \\
% \bottomrule
% \end{tabular}
\small\setlength{\tabcolsep}{1.7pt}
\begin{tabular}{lcccc}
\toprule
Models\hspace{-2mm} & English & Same local & Other local & Up.\,Sorbian \\
\midrule
Large & \textbf{4.97}\,{\tiny [4.97, 4.98]} & \textbf{4.97}\,{\tiny [4.96, 4.97]} & \textbf{4.92}\,{\tiny [4.90, 4.93]} & \textbf{4.85}\,{\tiny [4.83, 4.87]} \\
Med. & 4.94\,{\tiny [4.94, 4.95]} & 4.91\,{\tiny [4.90, 4.92]} & 4.87\,{\tiny [4.86, 4.88]} & 4.29\,{\tiny [4.26, 4.31]} \\
Small & 4.70\,{\tiny [4.69, 4.72]} & 4.45\,{\tiny [4.43, 4.47]} & 4.44\,{\tiny [4.41, 4.46]} & 1.43\,{\tiny [1.41, 1.45]} \\
\midrule
All & 4.87\,{\tiny [4.86, 4.88]} & 4.77\,{\tiny [4.76, 4.78]} & 4.74\,{\tiny [4.72, 4.75]} & 3.46\,{\tiny [3.44, 3.47]} \\
\bottomrule
\end{tabular}
\caption{Mean Kimi K3 faithfulness by prompt-language condition and model-size
group. Brackets show 95\% confidence intervals.\textsuperscript{\ref{fn:bootstrap}}}
\label{tab:language-condition-results}
\end{table}

\subsection{Factual, Fictional and Counterfactual}
\label{sec:variant-contrasts}

We now compare the three input variants directly. Of the 1{,}023 factual
items, 912 also have a fictional and a counterfactual counterpart, so we
restrict this analysis to those. Under Kimi as the judge, the score ordering is FA > FI > CFA in all model groups and overall, and
95\% \emph{pairwise} confidence intervals from resampling dataset--item clusters 10{,}000
times exclude zero for every contrast. The effect is nonetheless tiny (cf.~Table~\ref{tab:faithfulness-variants}): the CFA and FI faithfulness penalties are
$-0.049$ and $-0.030$ on a 1--5 scale. The same ordering holds for fluency in
every model-size group, and the counterfactual rewrite costs marginally more
fluency than faithfulness ($-0.071$ versus $-0.049$). DeepSeek produces
substantially larger CFA--FA differences in both faithfulness ($-0.177$) and
fluency ($-0.198$), but we regard Kimi as the more reliable judge.

%Table~\ref{tab:faithfulness-variants} splits the paired variant effect by model
%size: the FA $>$ FI $>$ CFA ordering holds in every size group, but Kimi
%compresses the overall means to 4.49, 4.46 and 4.44, and the large-group means
%are closely clustered, with overlapping marginal intervals.

\begin{table}[t]
\centering
\small
\begin{tabular}{lccc}
\toprule
Models & FA & FI & CFA \\
\midrule
Large   & \textbf{4.94} {\tiny [4.93, 4.95]} & \textbf{4.92} {\tiny [4.91, 4.94]} & \textbf{4.92} {\tiny [4.90, 4.93]} \\
Medium  & 4.78 {\tiny [4.76, 4.79]} & 4.76 {\tiny [4.74, 4.77]} & 4.73 {\tiny [4.71, 4.75]} \\
Small   & 3.80 {\tiny [3.78, 3.82]} & 3.75 {\tiny [3.73, 3.77]} & 3.73 {\tiny [3.71, 3.76]} \\
\midrule
All & 4.49 {\tiny [4.48, 4.50]} & 4.46 {\tiny [4.44, 4.47]} & 4.44 {\tiny [4.42, 4.45]} \\
\bottomrule
\end{tabular}
\caption{Mean Kimi faithfulness by true data variant and model-size group on the
32{,}832 matched FA/FI/CFA outputs, pooling both datasets and all four prompt
languages. Brackets show 95\% confidence intervals.\textsuperscript{\ref{fn:bootstrap}}}
\label{tab:faithfulness-variants}
\end{table}

\paragraph{Judge robustness checks.}
The judge prompt does not say if an input is FA, FI or CFA.
Disclosing this raises DeepSeek's scores by $0.148$, and its agreement with
the human labels drops ($\kappa_w$ $0.592 \rightarrow 0.499$). Kimi shifts only by
$0.050$ ($\kappa_w$ $0.717 \rightarrow 0.709$), within rerun noise.
Asking either judge explicitly to classify the sentence it is scoring changes little: the judges  neither
recover the factuality condition reliably, nor penalise sentences they believe
counterfactual beyond what the annotators do.
Appendix~\ref{sec:judge-audit-details} has details on both checks.

\paragraph{Perceived factuality.}
Table~\ref{tab:faithfulness-predicted-class-CI} examines whether faithfulness
varies with how generation models themselves perceive the input, using classification vote shares as
soft assignments. 
With Kimi as the judge, the ordering differs across model groups, with only small models showing FA > FI > CFA strongly.
%Under Kimi, perceived CFA has a lower mean for the large and
%small groups. However, the large-group means differ by only 0.02 points, while
%the medium group does not follow the FA $>$ FI $>$ CFA ordering. DeepSeek
%places perceived CFA below FA and FI in all three groups,
DeepSeek again shows a larger and less trustworthy effect.
Additional details are given in Appendix~\ref{sec:additional-results}.

%Table~\ref{tab:faithfulness-true-perceived} relates perceived to true
%factuality: perceived CFA remains the lowest within each true variant, whereas
%the reverse conditioning is less stable. Appendix
%Table~\ref{tab:faithfulness-predicted-class-detailed} gives the per-model breakdown.

\begin{table}[t]
\centering
\small\setlength{\tabcolsep}{4pt}
\begin{tabular}{lccc}
\toprule
Models & Pred. FA & Pred. FI & Pred. CFA \\
\midrule
Large  & \textbf{4.94} {\tiny [4.93, 4.95]} & \textbf{4.93} {\tiny [4.91, 4.94]} & \textbf{4.92} {\tiny [4.91, 4.93]} \\
Medium & 4.72 {\tiny [4.71, 4.74]} & 4.80 {\tiny [4.78, 4.82]} & 4.80 {\tiny [4.79, 4.82]} \\
Small  & 3.88 {\tiny [3.86, 3.90]} & 3.64 {\tiny [3.62, 3.66]} & 3.44 {\tiny [3.40, 3.47]} \\
\bottomrule
\end{tabular}
\caption{Mean faithfulness scores (1--5) by model-size group and the
model's classification of the input RDF triples, soft-weighted by classification vote share.
Pred.\ denotes the predicted class. Scores are from Kimi K3; brackets show 95\%
confidence intervals.\textsuperscript{\ref{fn:bootstrap}}}
\label{tab:faithfulness-predicted-class-CI}
\end{table}

\paragraph{Judge--human gap reflects factuality.}
The LLM judge and humans score the same outputs, so their difference reflects judge
behaviour rather than generator quality. On the 648 human-annotated rows, the
Kimi--human gap shrinks from $+0.35$ on FA to $+0.25$ on FI to $+0.16$ on CFA (or from $+0.33$ to $+0.28$ and $+0.15$ if grouping by perceived classes). 
%, an FA-to-CFA change of $-0.194$.
%Grouping the same rows by
%the generating model's classification of its input triples instead of by the true variant 
%gives $+0.33$, $+0.28$ and $+0.15$.

\paragraph{Explicit conflict acknowledgement.}
\label{sec:conflict-ack-ablation}

As the base generation prompt does not explicitly mention conflicts with prior knowledge, we re-run the generation with an alternative prompt, instructing models to treat all triples as authoritative and avoid correction, hedging, refusal or disclaimers.
% reports the per-model effect.
The pooled change is only $-0.012$ on the 1--5 scale, with no effect on ordering (see Appendix Table~\ref{tab:prompt-ablation}).
%Confidence intervals exclude zero for five out of nine models and the pooled result.
88\% of all outputs get the same score and nearly half are even textually identical. 
%Acknowledgment-style prose remains almost absent.
The largest score changes only reflect formatting changes in Upper Sorbian (e.g., inclusion of `|' separators). 
%The instruction
%therefore changes formatting more than explicit conflict handling and does not
%mitigate the pooled effect.

\subsection{Error analysis}
To better understand the faithfulness scores, we group the judge's error annotations into six non-exclusive categories: use of incorrect language or copying input triples unchanged (Lang), hallucinating unsupported information (Hall), reversing relation directions (Rev), missing source facts (Miss), incorrect relations between correct entities (Rel), adding/substituting an incorrect entity (Ent). Detailed definitions and examples are in Appendix~\ref{sec:mistake-examples}.

Table~\ref{tab:faithfulness-error-types-predicted-size-all} reports Kimi's
structured error labels. We observe about twice as many hallucinations for FI (highest) and CFA
compared to FA data in both medium and small-sized models, whereas the results are generally stable for large models. No error category follows the
FA--FI--CFA ordering consistently across model sizes.

\begin{table}[t]
\centering
\small\setlength{\tabcolsep}{3pt}
\begin{tabular}{llrrrrrr}
\toprule
Models & Class & Lang & Hall & Rev & Miss & Rel & Ent \\
\midrule
\multirow{3}{*}{Large} &
FA & 0.7 & 0.6 & 0.4 & 0.3 & 0.7 & 0.6 \\
& FI & 0.6 & 0.6 & 0.7 & 0.4 & 1.3 & 0.6 \\
& CFA & 0.8 & 0.7 & 0.8 & 0.5 & 0.7 & 0.7\Bstrut \\
\hdashline[0.5pt/2pt]
\multirow{3}{*}{Medium} &
FA & 4.9 & 1.3 & 1.3 & 1.0 & 1.5 & 0.6\Tstrut \\
& FI & 1.7 & 2.4 & 1.8 & 1.7 & 2.2 & 0.9 \\
& CFA & 2.2 & 2.2 & 1.3 & 1.3 & 2.2 & 1.0\Bstrut \\
\hdashline[0.5pt/2pt]
\multirow{3}{*}{Small} &
FA & 20.6 & 3.6 & 2.9 & 9.7 & 5.2 & 2.1\Tstrut \\
& FI & 23.7 & 12.8 & 2.6 & 7.1 & 4.1 & 4.7 \\
& CFA & 34.6 & 7.2 & 2.9 & 8.1 & 6.1 & 3.4 \\
\bottomrule
\end{tabular}
\caption{Kimi K3 error-category incidence (\% outputs) by predicted class and
model size; categories are non-exclusive. The number of outputs in each row
varies with the model prediction distribution in Table~\ref{tab:model-results}.}
\label{tab:faithfulness-error-types-predicted-size-all}
\end{table}

However, some CFA errors explicitly reveal the conflict between the provided data and the LLMs' parametric knowledge: despite instructions to use only the input triples and preserve all values exactly, Llama4-Scout generated the historically accurate year 1620 when the Battle of White Mountain was given the counterfactual date 2013, and rejected 1848 as singer Karel Kryl's album release date because it preceded his birth. These examples show that LLMs may favour memorised knowledge over the input, giving plausible but unfaithful outputs.

\section{Conclusion}
Across nine open-weight models and four languages, larger generators are
generally more faithful and all models struggle most in Upper Sorbian. 
%These broad trends hold under both judges. 
However, we find little support for the context--memory-conflict story. 
On matched FA-CFA items, the CFA faithfulness penalty is $-0.049$ under our
primary Kimi K3 judge, about one percent of the scale, and the FA $>$ FI $>$ CFA
ordering it preserves is numerical rather than substantive.
The CFA penalty is in fact more pronounced for fluency than faithfulness.
%On matched items the
%counterfactual rewrite costs more fluency than faithfulness under both Kimi K3 and DeepSeek V4 Pro Preview (our first judge).
An explicit
instruction to treat contradictory triples as authoritative does not reduce the
penalty either. Validated against human labels, the judges are
themselves mildly kinder to factual inputs, by several times the residual gap.
We therefore read the headline context--memory-conflict effect in this setting
as small and not cleanly separable from evaluator bias and the general
difficulty of realizing atypical inputs.

The broader lesson concerns evaluation rather than generation. On exactly the
same outputs, DeepSeek V4 Pro Preview puts the counterfactual penalty at $-0.177$, three
to four times higher, which on its own would have supported a clean
context--memory-conflict claim about the generators. Nothing in that judge's
aggregate agreement with our annotators would have flagged the problem
($\kappa_w$ 0.555, 80.9\% of scores within one point): a judge can agree well
overall and still carry a bias aligned with the experimental condition, and that
bias is then easily read as a property of the generators. Manual annotation held
out from prompt tuning ($\kappa_w$ 0.350), and evaluating alternative judges, is what separates a property of
the models from a property of the measurement.

\section*{Acknowledgments}
This work was funded by the European Union (ERC, NG-NLG, 101039303), the Czech AI Factory (CZAI) project (10131474, EuroHPC JU), and project CZ.02.01.01/00/23\_020/0008518 of the Czech Ministry of Education.
We also acknowledge support of the Czech Ministry of Education through the e-INFRA CZ research infrastructure (MetaCentrum \& CERIT-SC, ID:90254).
%,  MetaCentrum and CERIT-SC computing infrastructures.
%for providing the computing infrastructure and services we used to evaluate the open-weight large language models in our experiments.
We thank the anonymous reviews for their helpful comments, especially motivating us to use a better judge.

\section*{Limitations}
The main limitations of our work we are aware of are the following:
\begin{itemize}
    \item None of the authors participating in manual annotation for judge tuning was an Upper Sorbian speaker. However, all were native or near-native speakers of Czech and Slovak and given the languages' similarity, they were able to check the factuality of Upper Sorbian texts.
    \item We tuned the LLM-judge prompt on the 540-row sample and report agreement on it. The sample is stratified partly by an earlier judge score and has one author label per row without independent adjudication, so those numbers may overestimate generalization. The disjoint 108-row sample, annotated after the prompt was frozen tries to fix this little bit.
    \item To speed up generation, we ran all of the models with reasoning off or with a low reasoning effort, which may alter their behaviour compared to higher reasoning modes.

\end{itemize}

% =====================================================================
% NOTES TO AUTHORS -- REMOVE BEFORE SUBMISSION
% =====================================================================
% \clearpage
% \section*{TODO --- notes to authors}
% \emph{Working notes; delete this section before submission.}

% \begin{enumerate}
%   \item \textbf{Settle on the main message.} Revise abstract, intro, conclusion.
%   \item \textbf{Revise the tables.} Several tables are dense with numbers
%     (Tables~\ref{tab:model-results}, \ref{tab:direct-judge-comparison},
%     \ref{tab:faithfulness-variants},
%     \ref{tab:faithfulness-predicted-class-CI}). Decide per table whether the
%     confidence intervals, the per-model rows, or the second judge belong in
%     the main body at all, and move the rest to the appendix. Or a figure may
%     replace at least one of them.

%     \item \textbf{Github repo}, Add the new code (not much, mainly the one doing tables), add new annotations, maybe other things

% \end{enumerate}

\bibliography{bibliography}

\appendix

\section{Data Creation Pipeline}

\begin{figure}[t]
    \centering
    \includegraphics[width=0.9\linewidth]{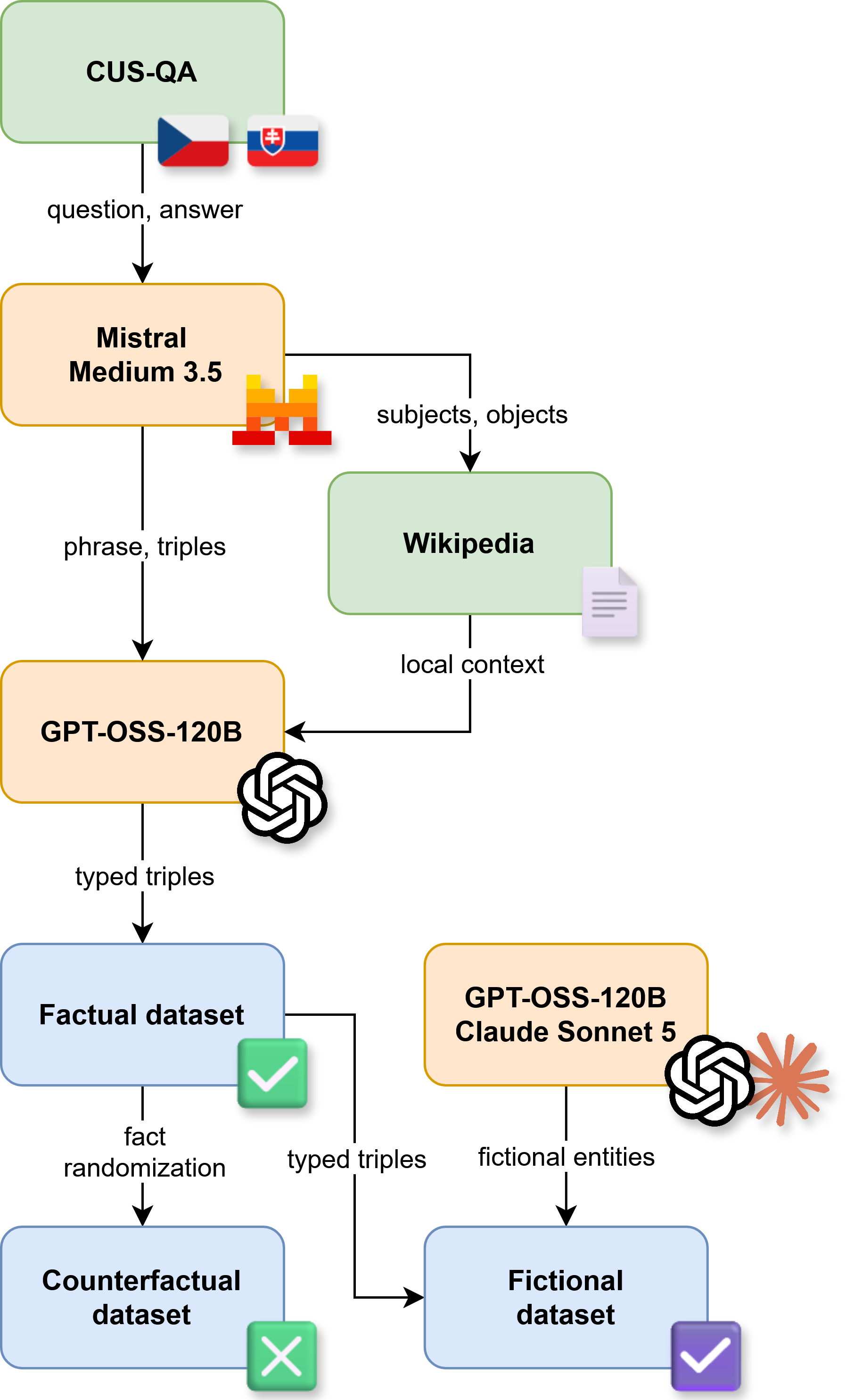}
    \caption{Pipeline to convert the CUS-QA dataset into RDF triples. First, we rewrite question–answer pairs into declarative phrases and decompose them into triples; then assign subject and object types using local Wikipedia context. We produce the CFA set by randomly swapping entities of the same type, and the FI set by injecting fictional LLM-generated entities.}
    \label{fig:cusqa}
\end{figure}

The schema for the data creation pipeline described in Section~\ref{sec:data} is shown in Figure~\ref{fig:cusqa}.

\section{Manual Annotation Details}
\label{sec:manual-annot-details}

\paragraph{Sampling strategy.}
Since the generator models are generally very capable, sampling uniformly would likely result in most examples scoring 5.
We therefore first ran the judging with an initial prompt, to gain some information about which triple groups were challenging in which languages.
Afterwards, we sampled the triple groups and the generated sentences in a way to uniformly cover every variant (FA, CFA, FI), every language (EN, CS, SK, HSB), three faithfulness score groups (score 1, scores 2--4, score 5), and three parameter-count strata (below 10B, 10--100B, and above 100B). These strata were used only to construct a diverse annotation sample and differ from the model-size groups used in the main analysis.
For each combination, we picked 5 random examples, giving us a total of $3 \times 4 \times 3 \times 3 \times 5 = 540$ examples to annotate. 
Within the broader faithfulness-score and model-size groups, we represented individual scores and models as evenly as possible, maximizing the representation of the least frequent one.
The variant, language and parameter-count strata are exactly balanced: each of the 36 combinations contributes 15 rows.
The faithfulness-score strata are not, because outputs scoring 1 are rare for the larger models outside Upper Sorbian, so many cells could not supply 5 of them and were backfilled from the other two strata.
The realized sample therefore contains 66 rows from the score-1 stratum, 218 from scores 2--4, and 256 from the score-5 stratum.
The 108-row held-out sample was drawn later with the same design and one example per combination, and is likewise balanced on variant, language and parameter count.

\paragraph{Annotation procedure.}
Before annotation, we hid the judge scores and identifiers of the generator models.
The four annotators first agreed on shared annotation guidelines.
Every unclear example encountered afterwards was discussed, and the agreed-upon decision was used to extend the guidelines.
Finally, we reviewed and updated the scores once more using the completed guidelines, obtaining the author reference labels used to optimize the LLM judge. Each row nevertheless has one final label rather than independent multi-annotator adjudication.

\section{Model Configuration Details}
\label{sec:model-config}

All models used BF16 precision except Qwen3.5-122B-A10B, which used FP8, and
gpt-oss-120b, which used MXFP4. We accessed Qwen3.5-122B-A10B, Gemma 4 31B,
gpt-oss-120b, and Llama 4 Scout through an OpenAI-compatible cloud service.\footnote{The quantization precision of cloud-hosted models reflects the deployment configuration available through the respective service.} 
We ran the remaining models locally via vLLM.
We used the same task-specific pipelines and prompt templates for all models.

All models were used with reasoning disabled when the API or backend supported
it. The only exception is gpt-oss-120b, where reasoning cannot be fully
turned off, so we set its reasoning level to low.

\section{Judge Robustness and Audit Details}
\label{sec:judge-audit-details}

\paragraph{DeepSeek language-condition results.}
Table~\ref{tab:deepseek-language-condition} is the direct robustness counterpart
to the primary Kimi results in Table~\ref{tab:language-condition-results}. The
ordering is identical, although Kimi is higher in every cell except the nearly
unchanged small-model Upper Sorbian result (1.43 versus 1.42).

\begin{table}[t]
\centering
\small
\begin{tabular}{lcccc}
\toprule
Models & EN & Same & Other & HSB \\
\midrule
Large & 4.89 & 4.87 & 4.83 & 4.55 \\
Medium & 4.85 & 4.80 & 4.72 & 4.00 \\
Small & 4.63 & 4.33 & 4.22 & 1.42 \\
\midrule
Overall & 4.78 & 4.66 & 4.57 & 3.26 \\
\bottomrule
\end{tabular}
\caption{Mean DeepSeek V4 Pro Preview faithfulness by prompt-language condition and
model-size group on the same outputs as Table~\ref{tab:language-condition-results}.}
\label{tab:deepseek-language-condition}
\end{table}

\paragraph{DeepSeek structured error labels.}
Table~\ref{tab:deepseek-error-types} shows that DeepSeek reports an
FA-to-FI-to-CFA rise in hallucination labels for every size group, whereas
Kimi yields a different pattern. Because the judges' structured labels are less stable than their scores,
we interpret the difference as judge sensitivity rather than gold error incidence.

\begin{table}[t]
\centering
\scriptsize\setlength{\tabcolsep}{2.5pt}
\begin{tabular}{llrrrrrr}
\toprule
Models & Class & Lang & Hall & Rev & Miss & Rel & Ent \\
\midrule
\multirow{3}{*}{Large} & FA & 1.6 & 3.2 & 0.6 & 0.1 & 0.6 & 1.7 \\
 & FI & 2.0 & 7.0 & 1.1 & 0.2 & 1.2 & 2.6 \\
 & CFA & 1.8 & 10.7 & 1.8 & 0.2 & 1.5 & 2.9 \\
\midrule
\multirow{3}{*}{Medium} & FA & 5.2 & 6.9 & 1.8 & 0.3 & 1.6 & 2.0 \\
 & FI & 3.1 & 9.1 & 3.1 & 0.5 & 1.9 & 3.0 \\
 & CFA & 4.3 & 15.2 & 2.7 & 0.5 & 2.7 & 2.9 \\
\midrule
\multirow{3}{*}{Small} & FA & 22.3 & 13.0 & 2.9 & 3.2 & 4.7 & 3.7 \\
 & FI & 21.9 & 15.7 & 3.2 & 2.4 & 2.8 & 5.7 \\
 & CFA & 34.3 & 15.9 & 4.1 & 3.1 & 5.0 & 5.1 \\
\bottomrule
\end{tabular}
\caption{DeepSeek V4 Pro Preview structured error-category incidence (\% of paired
outputs) by predicted class and model size. Categories are non-exclusive.}
\label{tab:deepseek-error-types}
\end{table}

\paragraph{Direct judge comparison.}
We compared Kimi and DeepSeek directly on both faithfulness and fluency. 
Table~\ref{tab:direct-judge-comparison} shows differences in faithfulness assessments (split per language and agreggated over all generating models), with Kimi generally assigning higher scores.
Table~\ref{tab:fluency-judge-agreement} then shows a similar breakdown for fluency; here, the agreement between judges is higher overall, with no clear trend in scores.

\begin{table}[t]
\centering
\small\setlength{\tabcolsep}{3pt}
\begin{tabular}{lrrrrrr}
\toprule
Condition & D & K & $\Delta$ & Exact & $\leq 1$ & $\kappa_w$ \\
\midrule
English & 4.78 & 4.87 & +0.09 & 87.7 & 96.4 & 0.474 \\
Same local & 4.66 & 4.77 & +0.11 & 83.5 & 94.6 & 0.554 \\
Other local & 4.57 & 4.74 & +0.16 & 80.7 & 92.0 & 0.451 \\
Upper Sorbian & 3.26 & 3.46 & +0.20 & 77.4 & 84.1 & 0.708 \\
\midrule
Overall & 4.32 & 4.46 & +0.14 & 82.3 & 91.8 & 0.717 \\
\bottomrule
\end{tabular}
\caption{Direct paired faithfulness comparison. D is DeepSeek V4 Pro Preview, K is
Kimi K3, $\Delta=\mathrm{K}-\mathrm{D}$, Exact and $\leq 1$ are the percentages
of outputs on which the two judges agree exactly and within one point, and
$\kappa_w$ is quadratic-weighted Cohen's kappa.}
\label{tab:direct-judge-comparison}
\end{table}

\begin{table}[t]
\centering
\small\setlength{\tabcolsep}{4pt}
\begin{tabular}{lrrrr}
\toprule
Condition & D & K & $\Delta$ & $\kappa_w$ \\
\midrule
English & 4.38 & 4.34 & -0.04 & 0.741 \\
Czech & 4.16 & 4.06 & -0.10 & 0.668 \\
Slovak & 4.02 & 3.93 & -0.09 & 0.698 \\
Upper Sorbian & 2.39 & 2.43 & +0.04 & 0.748 \\
\midrule
Overall & 3.74 & 3.69 & -0.05 & 0.805 \\
\bottomrule
\end{tabular}
\caption{Paired comparison of DeepSeek V4 Pro Preview (D) and Kimi K3 (K) fluency
scores on all 102{,}492 outputs. $\Delta=\mathrm{K}-\mathrm{D}$, and $\kappa_w$
is quadratic-weighted Cohen's kappa. The judges agree exactly on 62.8\% of
outputs and within one point on 90.8\%.}
\label{tab:fluency-judge-agreement}
\end{table}

\paragraph{Reasoning effort in the audit judges.}
On the same 540 rows, GPT-5.6 Sol high and low agree exactly on 87.8\% of scores and
within one point on 97.6\%. High reasoning modestly improves agreement with the
human labels while increasing observed reasoning tokens from 35{,}596 to
75{,}536. Agreement figures for all five judges are in
Table~\ref{tab:human-judge-agreement}.

\paragraph{Population-scale maximum disagreements.}
The 3{,}337 four-point gaps concentrate on target-language and raw-RDF-format
decisions, especially in Upper Sorbian. Examples include Kimi accepting raw
predicate names that DeepSeek treats as an untransformed triple, and Kimi
correctly rejecting Czech, English, Polish, or visibly corrupted mixed-script
grammar under another target-language prompt. These examples identify rubric
boundaries rather than ground-truth winners; the full sentences, triples,
scores, and judge comments are retained in the released disagreement CSV.

\begin{figure}[t]
\centering
\includegraphics[width=\columnwidth]{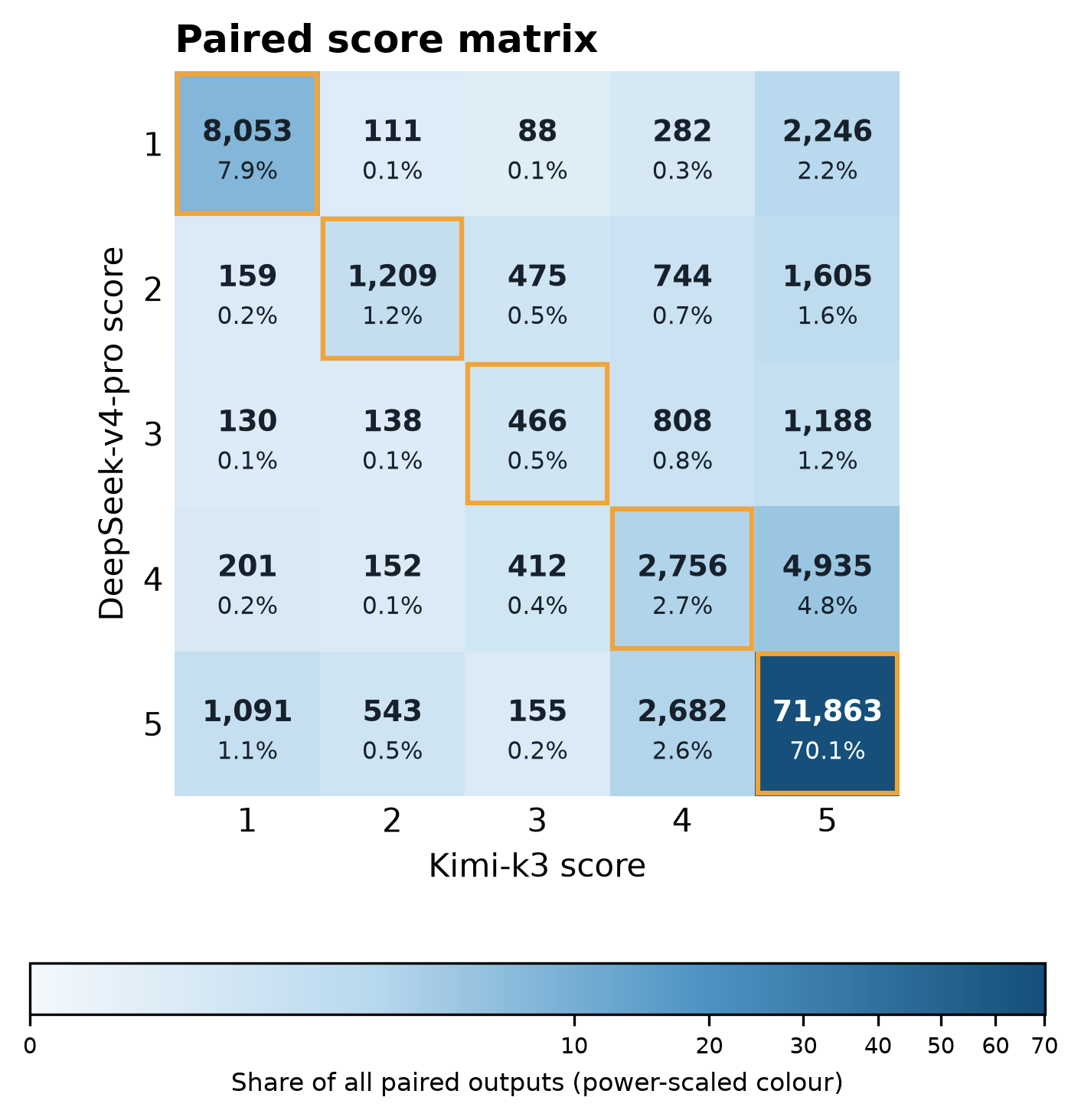}
\caption{Paired faithfulness-score matrix. Cells show counts and percentages
of all paired outputs; orange outlines mark exact agreement. Colour intensity
uses a power scale so that less frequent disagreements remain visible.}
\label{fig:judge-score-matrix}
\end{figure}

Upper Sorbian is one quarter of the data but contributes 44.5\%
of total absolute score difference and 79.8\% of the 3{,}337 four-point
(maximum) gaps. Of those gaps, 2{,}246 are DeepSeek 1 versus Kimi 5 and 1{,}091
run in the other direction. Figure~\ref{fig:judge-score-matrix} shows the paired
scores over the whole grid: 70.1\% of all outputs receive a 5 from both judges,
and the disagreement that remains is asymmetric, concentrated in the column
where Kimi awards 5.

\paragraph{Disclosing the class to the judge.}
The default judge prompt is blind: it never says whether the triples are
FA, FI or CFA. The disclosure run adds one line stating the condition, everything else is identical.
Because a judge rerun is not deterministic at the item level, we also ran the
blind prompt a second time under the same configuration. That \emph{replicate}
run is a noise floor, any disclosure effect smaller than it is not measurable.
Table~\ref{tab:judge-disclosure} reports both.

\begin{table}[t]
\centering
\small
\setlength{\tabcolsep}{4pt}
\begin{tabular}{lcccc}
\toprule
Run & Mean & Mean shift & Changed & $\kappa_w$ \\
\midrule
\multicolumn{5}{l}{\emph{DeepSeek}} \\
\quad blind      & 3.865 & ---                            & ---  & .592 \\
\quad replicate  & 3.850 & $-0.015$ {\tiny [-0.07, 0.04]} & 11.1\% & .568 \\
\quad disclosed  & 4.013 & $+0.148$ {\tiny [\phantom{-}0.05, 0.25]} & 23.5\% & .499 \\
\addlinespace
\multicolumn{5}{l}{\emph{Kimi}} \\
\quad blind      & 4.061 & ---                            & ---  & .717 \\
\quad replicate  & 4.063 & $+0.002$ {\tiny [-0.05, 0.05]} & 8.9\%  & .696 \\
\quad disclosed  & 4.111 & $+0.050$ {\tiny [-0.00, 0.11]} & 11.1\% & .709 \\
\bottomrule
\end{tabular}
\caption{Effect of disclosing the FA/FI/CFA class of the input to the judge, on
the 540-item annotation sample. \emph{replicate} is the blind prompt run twice
under the same configuration, giving the run-to-run noise floor. Mean shift is
against that judge's blind run, with 10{,}000-fold bootstrap percentile
intervals over paired per-item differences; \emph{Changed} is the share of items
whose score moves; $\kappa_w$ is quadratic-weighted agreement with the human
labels.}
\label{tab:judge-disclosure}
\end{table}

DeepSeek's disclosure effect exceeds its own noise floor on every measure: it
moves 23.5\% of items against a floor of 11.1\%, and costs $0.093$ of agreement
with the humans where a plain rerun costs $0.024$. The shift is upward, so the
disclosed judge is more lenient and \emph{less} accurate at once. Kimi's effect
does not exceed its noise floor on any measure --- its agreement degrades less under
disclosure ($-0.008$) than under a plain rerun ($-0.022$), and it moves 11.1\%
of items against a floor of 8.9\%. The two lifts also have inverted shapes:
DeepSeek's concentrates on CFA ($+0.25$, FA flat), Kimi's on FA ($+0.106$, CFA
flat at $+0.006$).

\paragraph{The class the judge perceives.}
Separately from the scoring runs, and without altering the judge prompt, we
asked each judge to label the generated sentences of the annotation sample as
FA, FI or CFA, given the same sentence and triples it sees when scoring. The
faithfulness scores below are the judges' ordinary scores for those
items, only the grouping is new.

Table~\ref{tab:judge-percept-confusion} shows
that neither judge can reliably predict the factuality, and that they fail in opposite
directions: DeepSeek defaults to FA, calling 44.7\% of counterfactual sentences
factual, while Kimi defaults to CFA, calling 44.4\% of factual sentences
counterfactual. 

\begin{table}[t]
\centering
\small\setlength{\tabcolsep}{5pt}
\begin{tabular}{lcccccc}
\toprule
& \multicolumn{3}{c}{DeepSeek} & \multicolumn{3}{c}{Kimi} \\
\cmidrule(lr){2-4}\cmidrule(lr){5-7}
True & FA & FI & CFA & FA & FI & CFA \\
\midrule
FA  & \textbf{.872} & .044 & .083 & \textbf{.544} & .011 & .444 \\
FI  & \textbf{.656} & .150 & .194 & .200 & .094 & \textbf{.706} \\
CFA & .447 & .084 & \textbf{.469} & .123 & .011 & \textbf{.866} \\
\bottomrule
\end{tabular}
\caption{How each judge classifies the generated sentence when shown the same
sentence and triples it scores, row-normalised over the true condition, 540
items. DeepSeek is biased towards FA and
Kimi towards CFA. Bold marks each judge's
most frequent label per row.}
\label{tab:judge-percept-confusion}
\end{table}

Table~\ref{tab:judge-percept-score} groups each judge's scores by its own
perceived class, with the human mean on the same items as an anchor. DeepSeek's
scores are flat across its perceived classes: items it calls CFA score $0.095$
below those it calls FA (CI includes zero) and the humans separate
the same items by $0.053$. Kimi does separate its perceived classes
($-0.328$), but by \emph{less} than the humans do on those same items
($-0.567$), so the gap reflects a real quality difference that Kimi
under-weights rather than a penalty for perceived counterfactuality. In neither
case does believing a sentence to be counterfactual make a judge harsher than a
human on that sentence.

\begin{table}[t]
\centering
\small
\begin{tabular}{llrcc}
\toprule
Judge & Percept & $n$ & Score & Human \\
\midrule
\multirow{3}{*}{DeepSeek}
 & FA  & 355 & 3.789 & 3.992 \\
 & FI  & \phantom{0}50 & 3.480 & 3.435 \\
 & CFA & 134 & 3.694 & 3.939 \\
\midrule
\multirow{3}{*}{Kimi}
 & FA  & 156 & 4.378 & 4.333 \\
 & FI  & \phantom{0}21 & 3.762 & 2.750 \\
 & CFA & 362 & 4.050 & 3.766 \\
\bottomrule
\end{tabular}
\caption{Mean score by the judge's own perceived class of the sentence, with
the human mean over the annotated subset of the same items.}
\label{tab:judge-percept-score}
\end{table}

\section{Additional Results}
\label{sec:additional-results}

This section complements the aggregate results in the main paper with analyses
by input size, uncertainty estimates for individual models, and detailed
results based on the ground-truth and perceived classes.

% \begin{figure}[t]
% \centering
% \includegraphics[width=\columnwidth]{fluency_vs_faithfulness.png}
% \caption{Mean faithfulness and fluency by prompt language on outputs scored in
% both dimensions by both judges. D is DeepSeek V4 Pro Preview and K is Kimi K3.}
% \label{fig:fluency-vs-faithfulness}
% \end{figure}

\paragraph{Joint true/perceived results.}
Table~\ref{tab:faithfulness-true-perceived} gives the primary Kimi results
behind the main-text summary.

\begin{table}[t]
\centering
\small
\setlength{\tabcolsep}{4pt}
\begin{tabular}{lccc}
\toprule
True class & Pred.\ FA & Pred.\ FI & Pred.\ CFA \\
\midrule
FA
  & 4.47 & 4.47 & 4.34 \\
  & {\scriptsize [4.45, 4.48]}
  & {\scriptsize [4.43, 4.51]}
  & {\scriptsize [4.30, 4.39]} \\
\midrule
FI
  & 4.43 & 4.47 & 4.36 \\
  & {\scriptsize [4.41, 4.44]}
  & {\scriptsize [4.43, 4.51]}
  & {\scriptsize [4.33, 4.39]} \\
\midrule
CFA
  & 4.42 & 4.46 & 4.35 \\
  & {\scriptsize [4.40, 4.44]}
  & {\scriptsize [4.42, 4.50]}
  & {\scriptsize [4.32, 4.38]} \\
\bottomrule
\end{tabular}
\caption{Mean Kimi K3 faithfulness by ground-truth and model-perceived input
class. Scores are macro-averaged across models and soft-weighted by the five
classification votes; brackets show 95\% confidence intervals.}
\label{tab:faithfulness-true-perceived}
\end{table}

\paragraph{DeepSeek joint true/perceived results.}
For direct comparison with the primary Kimi table, DeepSeek gives 4.41, 4.42,
and 4.22 for true FA across perceived FA/FI/CFA; 4.30, 4.35, and 4.14 for true
FI; and 4.26, 4.30, and 4.10 for true CFA. Unlike Kimi, true CFA is lowest
within all three perceived columns, although the differences are small.

\paragraph{Faithfulness by number of input triples.}
Table~\ref{tab:faithfulness-triples} gives the per-model results behind the
aggregate trend reported in the main paper. Scores generally decrease with the
number of triples under both judges. DeepSeek is monotonic for every model;
Kimi is nearly flat for GPT OSS 120B and rises from one to 2--3 triples for
Gemma4 E2B.

\begin{table}[t]
\centering
\small\setlength{\tabcolsep}{4pt}
\begin{tabular}{llcccc}
\toprule
Model & Judge & 1 & 2--3 & 4--9 & Overall \\
\midrule
\multirow{2}{*}{Qwen3.5 122B} & D & 4.94 & 4.72 & 4.60 & 4.77 \\
 & K & 4.95 & 4.89 & 4.81 & 4.90 \\
\multirow{2}{*}{GPT OSS 120B} & D & 4.95 & 4.75 & 4.66 & 4.80 \\
 & K & 4.95 & 4.96 & 4.96 & 4.96 \\
\hdashline[0.5pt/2pt]
\multirow{2}{*}{Gemma4 31B} & D & 4.92 & 4.64 & 4.52 & 4.71 \\
 & K & 4.94 & 4.93 & 4.93 & 4.94 \\
\multirow{2}{*}{Llama4 17B} & D & 4.84 & 4.57 & 4.45 & 4.64 \\
 & K & 4.80 & 4.76 & 4.76 & 4.77 \\
\multirow{2}{*}{Qwen3.5 9B} & D & 4.90 & 4.60 & 4.48 & 4.67 \\
 & K & 4.93 & 4.82 & 4.74 & 4.84 \\
\multirow{2}{*}{Gemma4 E4B} & D & 4.73 & 4.24 & 4.00 & 4.35 \\
 & K & 4.67 & 4.41 & 4.21 & 4.46 \\
\hdashline[0.5pt/2pt]
\multirow{2}{*}{Gemma4 E2B} & D & 4.22 & 3.76 & 3.59 & 3.88 \\
 & K & 4.12 & 4.16 & 4.05 & 4.12 \\
\multirow{2}{*}{Tiny Aya} & D & 3.80 & 3.55 & 3.50 & 3.62 \\
 & K & 3.73 & 3.64 & 3.57 & 3.66 \\
\multirow{2}{*}{Qwen3 1.7B} & D & 3.78 & 3.35 & 3.12 & 3.45 \\
 & K & 3.69 & 3.44 & 3.25 & 3.49 \\
\midrule
\multirow{2}{*}{All models} & D & 4.56 & 4.24 & 4.10 & 4.32 \\
 & K & 4.53 & 4.45 & 4.36 & 4.46 \\
\bottomrule
\end{tabular}
\caption{Average faithfulness by number of input triples. D and K denote
DeepSeek V4 Pro Preview and Kimi K3. Values pool both datasets, all prompt languages,
and all three variants.}
\label{tab:faithfulness-triples}
\end{table}

\paragraph{Confidence intervals for overall results.}
Table~\ref{tab:model-results-ci} reports uncertainty estimates for both judges
on the complete data. The intervals are narrow throughout, so the per-model
orderings in Table~\ref{tab:model-results} are not sampling noise. Faithfulness
broadly tracks model size under both judges, whereas the fluency ordering is
less regular and ranks Llama4 Scout and Qwen3.5 9B close to or above both large
models.

\begin{table}[t]
\centering
\scriptsize\setlength{\tabcolsep}{0.8pt}
\begin{tabular}{lcccc}
\toprule
& \multicolumn{2}{c}{Faithfulness} & \multicolumn{2}{c}{Fluency} \\
\cmidrule(r){2-3}\cmidrule(l){4-5}
Model & Kimi & DeepSeek & Kimi & DeepSeek \\
\midrule
\texttt{Qwen3.5-122B} & 4.90 {\tiny [4.88, 4.91]} & 4.77 {\tiny [4.75, 4.79]} & \textbf{4.15} {\tiny [4.12, 4.17]} & 4.16 {\tiny [4.14, 4.19]} \\
\texttt{gpt-oss-120b} & \textbf{4.96} {\tiny [4.95, 4.96]} & \textbf{4.80} {\tiny [4.78, 4.81]} & 3.98 {\tiny [3.96, 4.00]} & 3.97 {\tiny [3.95, 4.00]} \\
\hdashline[0.5pt/2pt]
\texttt{Gemma4-31B} & 4.94 {\tiny [4.93, 4.94]} & 4.71 {\tiny [4.69, 4.73]} & 3.76 {\tiny [3.74, 3.78]} & 3.85 {\tiny [3.82, 3.87]} \\
\texttt{Llama4-Scout} & 4.77 {\tiny [4.76, 4.79]} & 4.64 {\tiny [4.61, 4.66]} & 4.09 {\tiny [4.07, 4.11]} & \textbf{4.19} {\tiny [4.17, 4.21]} \\
\texttt{Qwen3.5-9B} & 4.84 {\tiny [4.83, 4.86]} & 4.67 {\tiny [4.65, 4.69]} & 4.13 {\tiny [4.11, 4.16]} & 4.18 {\tiny [4.16, 4.21]} \\
\texttt{Gemma4-E4B} & 4.46 {\tiny [4.44, 4.48]} & 4.35 {\tiny [4.33, 4.38]} & 3.49 {\tiny [3.47, 3.51]} & 3.61 {\tiny [3.58, 3.63]} \\
\hdashline[0.5pt/2pt]
\texttt{Gemma4-E2B} & 4.12 {\tiny [4.10, 4.15]} & 3.88 {\tiny [3.85, 3.91]} & 2.84 {\tiny [2.82, 2.86]} & 2.89 {\tiny [2.86, 2.91]} \\
\texttt{Tiny-Aya} & 3.66 {\tiny [3.64, 3.67]} & 3.62 {\tiny [3.60, 3.64]} & 3.74 {\tiny [3.72, 3.75]} & 3.75 {\tiny [3.73, 3.76]} \\
\texttt{Qwen3-1.7B} & 3.49 {\tiny [3.47, 3.51]} & 3.45 {\tiny [3.42, 3.47]} & 3.00 {\tiny [2.98, 3.03]} & 3.04 {\tiny [3.01, 3.06]} \\
\bottomrule
\end{tabular}
\caption{Mean Kimi K3 and DeepSeek V4 Pro Preview faithfulness and fluency scores by
generator. Brackets show 95\% confidence intervals.}
\label{tab:model-results-ci}
\end{table}

\paragraph{Per-model results by perceived class.}
Table~\ref{tab:faithfulness-predicted-class-detailed} expands the model-size
results in Table~\ref{tab:faithfulness-predicted-class-CI} into individual
models, with confidence intervals in place of the effective sample sizes we
previously used to signal which cells are thin. Under Kimi, five of nine models
(Qwen3.5 122B, Llama4 Scout, Qwen3.5 9B, Gemma4 E4B and Qwen3 1.7B) have a
Pred.\ FA advantage over Pred.\ CFA with disjoint intervals, marginally so for
Qwen3.5 9B. The differences
are much smaller than under DeepSeek, and two models reverse the direction:
Gemma4 E2B scores 4.09 versus 4.54 and Tiny Aya 3.17 versus 3.45, again with
disjoint intervals; GPT OSS 120B and Gemma4 31B overlap. Under DeepSeek the
FA-over-CFA direction holds for seven of nine models, with Gemma4 E2B and Tiny
Aya again reversed.

\begin{table}[t]
\centering
\scriptsize\setlength{\tabcolsep}{2pt}
\begin{tabular}{llccc}
\toprule
Model & J & Pred. FA & Pred. FI & Pred. CFA \\
\midrule
\multirow{2}{*}{Qwen3.5 122B} & D & 4.86 {\tiny [4.83, 4.87]} & 4.74 {\tiny [4.70, 4.79]} & 4.71 {\tiny [4.68, 4.73]} \\
 & K & 4.92 {\tiny [4.91, 4.94]} & 4.86 {\tiny [4.83, 4.90]} & 4.88 {\tiny [4.86, 4.90]} \\
\multirow{2}{*}{GPT OSS 120B} & D & 4.85 {\tiny [4.83, 4.87]} & 4.81 {\tiny [4.78, 4.84]} & 4.75 {\tiny [4.73, 4.78]} \\
 & K & 4.96 {\tiny [4.95, 4.97]} & 4.97 {\tiny [4.95, 4.98]} & 4.95 {\tiny [4.94, 4.96]} \\
\hdashline[0.5pt/2pt]
\multirow{2}{*}{Gemma4 31B} & D & 4.77 {\tiny [4.75, 4.79]} & 4.64 {\tiny [4.55, 4.72]} & 4.63 {\tiny [4.60, 4.67]} \\
 & K & 4.94 {\tiny [4.93, 4.95]} & 4.88 {\tiny [4.82, 4.92]} & 4.93 {\tiny [4.92, 4.95]} \\
\multirow{2}{*}{Llama4 17B} & D & 4.76 {\tiny [4.74, 4.78]} & 4.69 {\tiny [4.64, 4.73]} & 4.41 {\tiny [4.37, 4.45]} \\
 & K & 4.82 {\tiny [4.80, 4.83]} & 4.82 {\tiny [4.79, 4.85]} & 4.68 {\tiny [4.65, 4.71]} \\
\multirow{2}{*}{Qwen3.5 9B} & D & 4.73 {\tiny [4.71, 4.75]} & 4.65 {\tiny [4.62, 4.68]} & 4.59 {\tiny [4.56, 4.62]} \\
 & K & 4.86 {\tiny [4.85, 4.87]} & 4.82 {\tiny [4.80, 4.84]} & 4.83 {\tiny [4.80, 4.85]} \\
\multirow{2}{*}{Gemma4 E4B} & D & 4.36 {\tiny [4.34, 4.39]} & 4.35 {\tiny [4.24, 4.46]} & 4.20 {\tiny [4.09, 4.30]} \\
 & K & 4.46 {\tiny [4.44, 4.48]} & 4.52 {\tiny [4.43, 4.60]} & 4.34 {\tiny [4.24, 4.43]} \\
\hdashline[0.5pt/2pt]
\multirow{2}{*}{Gemma4 E2B} & D & 3.86 {\tiny [3.83, 3.88]} & 4.47 {\tiny [4.29, 4.63]} & 4.13 {\tiny [4.02, 4.24]} \\
 & K & 4.09 {\tiny [4.07, 4.12]} & 4.51 {\tiny [4.35, 4.66]} & 4.54 {\tiny [4.46, 4.63]} \\
\multirow{2}{*}{Tiny Aya} & D & 3.18 {\tiny [3.12, 3.23]} & 3.79 {\tiny [3.76, 3.81]} & 3.41 {\tiny [3.37, 3.45]} \\
 & K & 3.17 {\tiny [3.12, 3.22]} & 3.82 {\tiny [3.80, 3.84]} & 3.45 {\tiny [3.42, 3.49]} \\
\multirow{2}{*}{Qwen3 1.7B} & D & 3.67 {\tiny [3.65, 3.70]} & 3.04 {\tiny [2.99, 3.10]} & 2.48 {\tiny [2.41, 2.57]} \\
 & K & 3.74 {\tiny [3.71, 3.76]} & 2.98 {\tiny [2.93, 3.02]} & 2.58 {\tiny [2.50, 2.65]} \\
\bottomrule
\end{tabular}
\caption{Mean faithfulness grouped by the generator's self-classification and
soft-weighted by its five classification votes. Brackets show 95\% confidence
intervals; J is the judge: D is DeepSeek V4 Pro Preview and K is Kimi K3.}
\label{tab:faithfulness-predicted-class-detailed}
\end{table}

\paragraph{Significance of perceived-class differences.}
Under DeepSeek, perceived CFA is more difficult in every model-size group. For
large, medium, and small models, respectively, the
Bonferroni-adjusted 95\% family-wise confidence intervals for Pred.\ FA $-$
Pred.\ CFA were [0.09, 0.15], [0.04, 0.11], and [0.35, 0.47]; the corresponding
intervals for Pred.\ FI $-$ Pred.\ CFA were [0.01, 0.10], [0.05, 0.14], and
[0.25, 0.37]. Under Kimi, the medium-group direction differs and the
large-group differences shrink to near-ceiling margins. We therefore interpret
the DeepSeek intervals as judge-specific robustness results rather than the
primary conclusion.

\paragraph{A proprietary generator as a reference point.}
The nine evaluated models are all open-weight. To place their scores on a
scale, we additionally ran the full generation and classification pipeline with
DeepSeek V4 Pro Preview as a \emph{generator} and judged its outputs with Kimi K3.
Table~\ref{tab:deepseek-generator} gives the result. It is excluded from every
other table in this paper because we have only Kimi scores for it: we avoid
self-evaluation, so DeepSeek does not judge its own outputs.

Two observations follow. First, its mean faithfulness of 4.92 is comparable to
the best open-weight models, and it holds that level in Upper Sorbian (4.93,
slightly above the 4.85 of the large open-weight group and far above the 1.43
of the small one). The gap to the small models isolates target-language
competence rather than faithfulness as the binding constraint at the low end. Second, its
spread across the three data variants is 0.03 points, smaller than any
open-weight model's, while its classification accuracy of 49.3\% is below that
of gpt-oss-120b --- so recognizing counterfactual input and being derailed by
it are clearly separable abilities.

\begin{table}[ht]
\centering
\small
\begin{tabular}{llc}
\toprule
& Condition & Faithfulness \\
\midrule
& Overall & 4.92 {\scriptsize [4.91, 4.93]} \\
\midrule
\multirow{3}{*}{Variant}
 & FA & 4.94 {\scriptsize [4.92, 4.95]} \\
 & FI & 4.92 {\scriptsize [4.90, 4.93]} \\
 & CFA & 4.91 {\scriptsize [4.89, 4.92]} \\
\midrule
\multirow{4}{*}{Language}
 & EN & 4.98 {\scriptsize [4.97, 4.98]} \\
 & CS & 4.84 {\scriptsize [4.81, 4.86]} \\
 & SK & 4.94 {\scriptsize [4.93, 4.95]} \\
 & HSB & 4.93 {\scriptsize [4.91, 4.94]} \\
\bottomrule
\end{tabular}
\caption{DeepSeek V4 Pro Preview used as a \emph{generator} over all 11{,}388 inputs,
judged by Kimi K3 only. Brackets show 95\% confidence intervals. Its
majority classification accuracy is 49.3\% with 81.9\% unanimity, predicting
FA/FI/CFA on 69/18/13\% of inputs.}
\label{tab:deepseek-generator}
\end{table}

\paragraph{Conflict-Acknowledgement Generation Ablation}

Table~\ref{tab:prompt-ablation} shows results comparing the default prompt (Base) with an alternative prompt explicitly acknowledging potential conflicts with parametric knowledge (Ack.; see Appendix~\ref{app:conflict-ack-prompt}).

\begin{table}[t]
\centering
\small\setlength{\tabcolsep}{3pt}
\begin{tabular}{lrrrrr}
\toprule
Model & Base & Ack. & $\Delta$ & Sco.\,A. & Idnt.\,O. \\
\midrule
Qwen3.5 122B & 4.895 & 4.878 & -0.018 & 93.3\% & 37.4\% \\
GPT OSS 120B & 4.958 & 4.925 & -0.033 & 95.1\% & 23.4\%\Bstrut \\
\hdashline[0.5pt/2pt]
Gemma4 31B & 4.935 & 4.932 & -0.004 & 96.4\% & 64.5\%\Tstrut \\
Llama4 17B & 4.773 & 4.586 & -0.188 & 82.7\% & 26.2\% \\
Qwen3.5 9B & 4.843 & 4.849 & +0.006 & 91.9\% & 58.3\% \\
Gemma4 E4B & 4.459 & 4.473 & +0.014 & 89.5\% & 59.0\%\Bstrut \\
\hdashline[0.5pt/2pt]
Gemma4 E2B & 4.125 & 4.245 & +0.121 & 82.1\% & 52.8\%\Tstrut \\
Tiny Aya & 3.656 & 3.662 & +0.006 & 86.2\% & 48.2\% \\
Qwen3 1.7B & 3.486 & 3.470 & -0.017 & 82.5\% & 57.4\% \\
\midrule
Overall & 4.459 & 4.446 & -0.012 & 88.8\% & 47.5\% \\
\bottomrule
\end{tabular}
\caption{Conflict-acknowledgment generation-prompt ablation judged by Kimi K3.
$\Delta$ = acknowledgment minus baseline; Sco.\,A. = score agreement; Idnt.\,O.
= percentage of identical generations.}
\label{tab:prompt-ablation}
\end{table}

\paragraph{Per-language mistakes analysis.}

Table~\ref{tab:faithfulness-error-types-language} compares the error
categories defined in Section~\ref{sec:mistake-examples} across prompt-language
conditions under both judges, on the same explicit-label mapping as
Tables~\ref{tab:faithfulness-error-types-predicted-size-all}
and~\ref{tab:deepseek-error-types}, using the prompt-language conditions defined
in Section~\ref{sec:generation-results}. Each condition has the same number of
outputs.

\begin{table}[ht]
\centering
\scriptsize\setlength{\tabcolsep}{2.5pt}
\begin{tabular}{lrrrrrr}
\toprule
Condition & Lang & Hall & Rev & Miss & Rel & Ent \\
\midrule
EN & 0.1 / 0.1 & 11.1 / 3.1 & 1.8 / 1.4 & 0.5 / 2.1 & 0.9 / 0.9 & 2.4 / 1.9 \\
Same & 0.7 / 0.4 & 11.0 / 4.0 & 2.8 / 2.1 & 1.6 / 5.3 & 3.2 / 3.5 & 3.3 / 1.6 \\
Other & 2.5 / 1.3 & 12.6 / 4.2 & 2.8 / 2.1 & 1.6 / 5.2 & 3.1 / 3.4 & 3.8 / 1.5 \\
HSB & 37.6 / 36.2 & 7.2 / 1.5 & 1.6 / 1.0 & 1.1 / 1.4 & 2.5 / 2.5 & 2.8 / 1.0 \\
\bottomrule
\end{tabular}
\caption{Error-category incidence (\% of paired outputs) by prompt-language
condition. Each cell is DeepSeek / Kimi; categories are non-exclusive. Each
condition has 25{,}623 outputs.}
\label{tab:faithfulness-error-types-language}
\end{table}

The main difference is the high incidence of language or transformation errors
with HSB prompts, and it is the one error signal on which the two judges nearly
agree: 37.6\% of HSB outputs contain such errors under DeepSeek and 36.2\% under
Kimi, against 0.1\% for EN under both. Small models often failed to produce
a sentence in Upper Sorbian, resulting in faithfulness and fluency scores near
the bottom of the scale, as shown earlier in
Table~\ref{tab:language-condition-results}. The other categories are far less
stable: Kimi reports at most a third of DeepSeek's hallucination rate in every
condition, while reporting three to four times as many omissions outside Upper
Sorbian. This is the same judge sensitivity that the main text reports by
predicted class.

\section{Experimental Prompt Templates}
\label{sec:appendix-prompts}
This appendix provides the English versions of the prompt templates used in the
experiments. The classification and data-to-text generation prompts were
used in four prompt languages: English, Czech, Slovak, and Upper
Sorbian. The language-specific versions differ only in the language of the
instructions and output requirements. Braced expressions denote values filled
in for each example, such as the number of triples, the data category, the
target language, or the generated sentence.

\subsection{Classification Prompt}

\begin{Verbatim}[breaklines=true,breakanywhere=true,fontsize=\scriptsize,frame=single]
Classify a set of RDF triples as exactly one of these labels:
- FA (factual): the information is true and matches real-world knowledge.
- CFA (counterfactual): the entities are real, but one or more stated relationships or values are false or altered.
- FI (fictional): the entities themselves are fictional or do not exist in the real world.

Use only these three labels. Classify the whole set with one label, not each triple separately. You may reason internally if needed, but do not put reasoning in the final answer.

Below are {size} RDF triples from the thematic domain '{domain}'.

{triples_str}

Output only the label: FA, CFA, or FI.
\end{Verbatim}

\subsection{Data-to-Text Generation Prompt}

\begin{Verbatim}[breaklines=true,breakanywhere=true,fontsize=\scriptsize,frame=single]
You will be given RDF triples and must express them as fluent English prose.
Each triple is in subject | predicate | object order.
Generate a single fluent English sentence (or two short ones if needed) that naturally expresses ALL of the given facts.

Strict requirements:
- Use ONLY the information given in the triples. Do NOT add any facts, details, attributes, or context that are not explicitly stated.
- Do NOT omit any fact from the triples.
- Do NOT infer, assume, embellish, or generalize beyond what the triples literally state.
- Preserve all numbers, dates, and values exactly as given.
- Translate any entity that is NOT a proper named entity into English (e.g. common nouns, roles, generic descriptions). Keep proper named entities (names of people, places, organisations, works, etc.) in their original form.
- ALWAYS translate the predicate into English and use it naturally in the sentence.

Output ONLY the sentence(s), nothing else.

Below are {size} RDF triples from the thematic domain '{domain}': {triples_str}
\end{Verbatim}

\subsection{Conflict-Acknowledgment Addition}
\label{app:conflict-ack-prompt}

For the ablation in Section~\ref{sec:conflict-ack-ablation}, we add the
following paragraph to the English generation prompt; equivalent Czech,
Slovak, and Upper Sorbian versions are used for the other target languages.

\begin{quote}
\small\itshape
Important: Some inputs are intentionally designed to contradict common
knowledge or your own prior knowledge. For this task, treat every provided
triple as authoritative, even if it appears false, implausible, or inconsistent
with what you know. Express the claims exactly as given; do not correct them,
replace them with known facts, hedge, refuse, or add a disclaimer.
\end{quote}

\subsection{Faithfulness-Judge Prompt}

\begin{Verbatim}[breaklines=true,breakanywhere=true,fontsize=\scriptsize,frame=single]
You are judging faithfulness in data-to-text generation. Decide whether the sentence is fully supported by the triples below. The triples are the complete and ONLY source of truth. Do not use outside knowledge. Do not explain your reasoning. Output the final JSON immediately.

STEP 0 — GATES (check these before judging content):
- Wrong language / wrong transform, set faithfulness_score = 1 with one incorrect_information entry (comment label: wrong_language or wrong_transform):
   - if the sentence's own grammar — its verbs, function words, and connecting phrases — is not written mostly in {target_language}. Judge this only on the words the sentence adds around the triple terms, never on wording copied from the triples. Be especially careful if target language is low-resource, such as upper serbian (hsb), as even a technically valid sentence in different language should be scored low. However, if only parts of the sentence are not in the target language, while most of it is, be a bit more benevolent, and follow instructions below.
   - Any subject or object wording taken from the triples — named entities (people, places, organizations, titles, works) as well as descriptive or common-noun terms — routinely remains in its original language inside a correctly written {target_language} sentence. This alone is NEVER wrong_language and never justifies score 1, even when such copied wording makes up most of the sentence's length. If a translatable term was translated, left untranslated or translated imprecisely, it is at most a fluency problem and never a reason for lowering faithfulness score.
   - if it is not a natural-language sentence (e.g., it keeps the exact | pipes from the source triple). If it just keeps the RDF order, or does not split the predicate, it is alright, continue with instructions below
- Degenerate repetition: if the sentence loops or repeats the same clause/phrase at least 10 times, the final score is at most 3 (lower if it violates additional rules for content faithfulness); add an entry with comment label: degenerate. For info_used on this entry, do NOT quote the repeated span verbatim -- write a short placeholder like "(repeated/looping text)" instead, then continue directly to faithfulness_score. Minor redundancy is acceptable.

CORE RULES:
- Treat every triple as true, even if it is nonsense in the real world.
- NEVER penalize implausibility of any kind. All of the following are FULLY FAITHFUL (score 5) if they match the triples:
  - self-referential statements (e.g., "X is located in X.")
  - impossible or reversed time spans (e.g., "from 2000 to 1900")
  - geographically, historically, or otherwise logically absurd facts
- Interpret each triple as subject | predicate | object, using the predicate label's actual semantics. Some labels define the object's role: X | broadcastedBy | Y means Y broadcast X; X | riverFlowingThrough | Y means river Y flows through X; X | hasToItsNorth | Y means Y is north of X.
- Reversed relation = the sentence swaps the SEMANTIC roles of subject and object, so the resulting claim asserts the opposite of what the triple states. Do not mark a relation reversed just because the sentence order differs: many of the target languages (e.g. Czech, Slovak) mark grammatical role through case endings and agreement rather than word position, so a subject can correctly appear after the object. Determine the role each entity plays from grammar/meaning in context, not from which one appears first, and only flag reverse when the asserted relationship itself is actually flipped.
- Paraphrase is fine: any wording with the same meaning counts as supported. In particular, X | country | Y is often fully supported by "X is located in Y", "X is in Y", "X lies in Y" and similar. Apply the same tolerance to other predicates.
- Name translation is fine: translating descriptive or common-noun names into {target_language} is acceptable, as are inflection, diacritics, and transliteration changes. Only penalize a name change if it alters the entity's identity or meaning.
- Do not penalize word order, function words, or stylistic differences when the meaning is preserved.
- Omission: if any information from any triple is missing from the sentence, count it as an omitted claim (info_used = "(missing)", correct_info = the omitted triple).
- Only place a claim in incorrect_information if it is unsupported, contradicted, reversed, negated, omitted, over-specific, or incorrectly transformed relative to the triples. NEVER place supported information there.

SCORING RUBRIC (after gates):
- 5 = fully faithful: every claim explicitly supported (paraphrase allowed), nothing unsupported, nothing omitted — even if nonsensical, self-referential, or implausible. If incorrect_information is [], the score MUST be 5.
- 4 = one minor bug: exactly one wrong predicate or object, one small unsupported claim, one omitted claim, minor translation bug or similar; everything else correct.
- 3 = mixed: several problems, or supported content mixed with a clear substantive error, or a degenerate/looping sentence (cap).
- 2 = reversed relation, negated/contradicted predicate, or multiple major problems.
- 1 = wrong language, kept original pipe-separated-RDF format, or the sentence is largely unsupported / contradicts the triples overall.

Return STRICT JSON with these keys only:
- incorrect_information: array of objects with keys info_used, correct_info, comment
  - info_used: the exact claim from the sentence that is wrong/unsupported ("(missing)" for omissions)
  - correct_info: the triple-backed correction or missing constraint
  - comment: brief reason, starting with one label if applicable: unsupported | wrong_object | wrong_predicate | reverse | negation | omission | over_specific | wrong_language | wrong_transform | repetition
- faithfulness_score: integer from 1 to 5

Mini examples:
1. Triple: Havlíčkův Brod | riverFlowingThrough | Hrubý Jeseník
   Sentence: The Hrubý Jeseník river flows through Havlíčkův Brod.
   Output idea: faithfulness_score 5, incorrect_information []

2. Target language: English
   Triples:
   - Nejsevernější národní park v České republice | hasName | budova hlavního nádraží v Praze
   - budova hlavního nádraží v Praze | isLocatedIn | Česká republika
   - budova hlavního nádraží v Praze | hasProperty | nejsevernější
   Sentence: The northernmost national park in the Czech Republic is named the main railway station building in Prague. The main railway station building in Prague is located in the Czech Republic and has the property of being the northernmost.
   Output idea: faithfulness_score 5, incorrect_information [] — every claim matches its triple exactly. It does not matter that a park being "named" a building is real-world nonsense; do not reinterpret the subject/object relationship to make it more sensible.

3. Target language: English
   Triples:
   - Československá automobilová doprava | providesService | autobusová doprava
   - Československá automobilová doprava | operatesIn | Československo
   Sentence: Československá automobilová doprava provides autobusová doprava in Československo.
   Output idea: faithfulness_score 5, incorrect_information [] — the sentence's own grammar ("provides ... in ...") is entirely English; the Czech subject/object wording is copied straight from the triples, which is never wrong_language, no matter how much of the sentence's length it takes.

4. Target language: English
   Triple: Kateřina Bobková-Valentová | isBrotherOf | Svatý Václav
   Sentence: ...Kateřina Bobková-Valentová is his brother...
   Output idea: faithfulness_score 5, incorrect_information [] — "X isBrotherOf Y" means X is the brother of Y, so "Kateřina Bobková-Valentová is his (Svatý Václav's) brother" states exactly that relationship. Do not flip the roles based on which name a pronoun like "his" seems to attach to in the surrounding context — check only whether the sentence's claim matches the triple's own subject/object direction. Only call this reverse if the sentence instead said Svatý Václav is Kateřina's brother.

5. Target language: Upper Sorbian (hsb)
   Triple: Zlatá kopa | originatesIn | Národný park Nízke Tatry
   Sentence: Zlatá kopa | originatesIn | Národný park Nízke Tatry
   Output idea: faithfulness_score 1, comment "wrong_transform" — the sentence is literally the raw triple, not a natural-language sentence in any language.

6. Target language: English
   Triples:
   - okres Beroun | namedAfter | těžba stříbrných rud
   - okres Beroun | namedAfter | tavení stříbra
   Sentence: The okres Beroun is named after both the mining of silver ores and the smelting of silver.
   Output idea: faithfulness_score 5, incorrect_information [] — "the mining of silver ores" and "the smelting of silver" are faithful English paraphrases of the triples' Czech object phrases, not unsupported additions. Do NOT require the sentence to reuse the triple's exact source wording — judge the meaning, not the surface form, and do not flag a term as unsupported merely because it was translated or phrased differently than the triple.

7. Target language: English
   Triples:
   - Najvýchodnejší slovenský národný park | name | Národný park Poloniny
   - Národný park Poloniny | location | Čierny Balog
   - Národný park Poloniny | position | najvýchodnejší
   Sentence: The most eastern Slovak national park, named Národný park Poloniny, is located in Čierny Balog and is the most eastern.
   Output idea: faithfulness_score 5, incorrect_information [] — attaching a "name" triple to its subject via apposition ("The most eastern Slovak national park, named X, ...") and then continuing with the subject's other facts is a normal way to combine several triples about the same entity into one sentence. This is NOT merging the subject and object into a single entity or an unsupported claim — every fact still traces to its own triple.

8. Target language: Slovak
   Triples:
   - První parostrojní železnice v českých zemích | hasName | Severní dráha císaře Ferdinanda
   - Severní dráha císaře Ferdinanda | isLocatedIn | Nové Sady
   Sentence: Severní dráha císaře Ferdinanda je názov prvej parostrojnej železnice v českých zemiach, ktorá sa nachádza v Nových Sadoch.
   Output idea: faithfulness_score 5, incorrect_information [] — the sentence's own grammar ("je názov", "ktorá sa nachádza v") is Slovak; "Severní dráha císaře Ferdinanda" and "První parostrojní železnice v českých zemích" are entity names copied straight from the triples, in their original Czech form. Copied triple wording never counts toward "is the sentence mostly in {target_language}" — not even when, as here, it is most of the sentence's visible text. This is NOT wrong_language.

Target language: {target_language}
Category: {category}
Sentence: {sentence}

Triples:
{modified_triples}
\end{Verbatim}

\subsection{Fluency-Judge Prompt}

\begin{Verbatim}[breaklines=true,breakanywhere=true,fontsize=\scriptsize,frame=single]
You are judging the linguistic quality of a single sentence written in a given target language.

Task: decide quickly how well the sentence is written in that language.
Judge only the language: grammar, inflection and agreement, word order, spelling and orthography, punctuation, word choice, and how natural it sounds to a native speaker.
Do not explain your reasoning.
Output the final JSON immediately.

Return STRICT JSON with these keys only:
- fluency_score: integer from 1 to 5
- fluency_comment: a brief explanation of the language quality, in English

Important rules:
- Ignore whether the sentence is true, plausible, counterfactual, or nonsensical. Content correctness is NOT your concern — only linguistic quality.
- A sentence can be perfectly fluent even if what it says is false or impossible.
- Example: "Praha je nejvyšší hora na Slovensku." is completely fluent Czech and must score 5, even though it is factually false.
- Do not reward or penalise a sentence for the amount of information it contains.
- Judge the sentence as {language}. If it is written in a different language, that is a fluency failure.
- Some entities may not be in {language}: proper named entities (names of people, places, organisations, works, etc.) may remain in {entity_language} rather than being translated. Do NOT treat such untranslated proper names as a fluency error — judge only the quality of the surrounding {language} text.
- The source lexical terms below are deduplicated subjects and objects from the source triples, with RDF underscores already replaced by spaces. Proper names may be preserved as written. Non-proper common nouns, adjectives, roles, and descriptions should normally be translated or naturally rendered in {language}; awkward untranslated common terms may lower fluency.

Anchors:
- 5 means flawless, natural, native-quality {language}: no grammatical, spelling, or word-choice issues
- 4 means good: fully understandable with at most a minor slip that a native speaker would barely notice
- 3 means acceptable: understandable, but with noticeable grammatical, spelling, or word-choice errors
- 2 means poor: frequent or serious errors that make the sentence awkward or hard to read
- 1 means broken: severely ungrammatical, garbled, or not really {language}

Language: {language}
Source lexical terms:
{lexical_terms}

Sentence: {sentence}
\end{Verbatim}

\section{Mistake Categories and Examples}
\label{sec:mistake-examples}

\paragraph{Category mapping.}
The mistake categories are derived from the structured
\texttt{incorrect\_information} entries returned by the faithfulness judges. The
same mapping is applied to both judges' labels; the examples given below are
taken from the DeepSeek V4 Pro Preview annotations.
We consider only outputs with a faithfulness score below 5 when assigning an
error type. The reported percentages in
Table~\ref{tab:faithfulness-error-types-predicted-size-all}, however, use all
judged outputs in the corresponding condition as the denominator. Thus, they
describe the incidence of each error type among all generations, rather than
its proportion among erroneous generations only.

The judge may identify several incorrect claims in one output. We map the
fine-grained labels attached to these claims into six broader categories:
\texttt{wrong\_language}, \texttt{wrong\_transform}, and degenerate repetition
are grouped as language or transformation errors (\textsc{Lang});
\texttt{unsupported}, \texttt{hallucination}, and
\texttt{over\_specific} as hallucination (\textsc{Hall});
\texttt{reverse} as a reversed relation (\textsc{Rev});
\texttt{omission} as missing information (\textsc{Miss});
\texttt{wrong\_predicate} as an incorrect relation (\textsc{Rel}); and
\texttt{wrong\_object} as an incorrect entity or value (\textsc{Ent}).
The categories are non-exclusive because one output may contain several
different mistakes. Rare contradiction annotations and entries without a
recognized label are not included in these six categories.

\paragraph{Examples.}
The following examples illustrate the distinction between the categories.
They are selected from large, medium, and small models rather than from a
single model-size group.

\begin{description}

\item[\textsc{Lang}: language or transformation error.]
This category includes outputs written in a language other than the requested
one, outputs that retain RDF predicates instead of forming a natural-language
sentence, and strongly degenerate repetitions. For example,
\texttt{Qwen3.5-122B} was asked to produce an English sentence from:

\begin{quote}\small
\textit{Triples:}
kraj Vysočina $\mid$ isA $\mid$ historická země;\\
kraj Vysočina $\mid$ location $\mid$ severovýchod České republiky

\textit{Output:}
Kraj Vysočina je historická země nacházející se na severovýchodě
České republiky.
\end{quote}

The content follows the triples, but the complete sentence is in Czech rather
than English. This is therefore a language error rather than a hallucinated
fact.

\item[\textsc{Hall}: hallucination.]
This category covers information not supported by the source triples. It also
includes cases where the model replaces a counterfactual fact with information
that is more consistent with its general knowledge. For example,
\texttt{Tiny-Aya} received the counterfactual triple:

\begin{quote}\small
\textit{Triple:}
Rakúsko $\mid$ shortName $\mid$ PIENAP

\textit{Output:}
Rakousko, oficiálně známé jako Rakouská republika, je země v srdci
Evropy.\\
\textit{[Austria, officially known as the Republic of Austria, is a country
in the heart of Europe.]}
\end{quote}

Instead of stating that \textit{PIENAP} is the short name of Austria, the model
ignores the counterfactual relation and produces a factual, general description
of Austria. DeepSeek therefore assigns this output a faithfulness score of 1.

\item[\textsc{Rev}: reversed relation.]
A reversed relation preserves the relevant entities and often the general
relation, but exchanges their semantic roles. For example,
\texttt{Gemma4-E2B} produced:

\begin{quote}\small
\textit{Triple:}
Rákosníček $\mid$ voicedBy $\mid$ Ferdinand d'Este

\textit{Output:}
Ferdinand d'Este voiced by Rákosníček.
\end{quote}

The triple states that Rákosníček is voiced by Ferdinand d'Este, while the
output makes Rákosníček the voice of Ferdinand d'Este.

\item[\textsc{Miss}: missing information.]
An omission occurs when an entity, relation, or complete source triple is not
verbalized. For example, \texttt{Qwen3.5-122B} received:

\begin{quote}\small
\textit{Triple:}
Národní divadlo v Praze $\mid$ foundationStoneLaidDate
$\mid$ 1868-01-01

\textit{Output:}
Kámen základní byl položen dne 1868-01-01.\\
\textit{[The foundation stone was laid on 1868-01-01.]}
\end{quote}

The date and event are expressed, but the output does not say that they refer
to the National Theatre in Prague. The source entity is therefore missing.

\item[\textsc{Rel}: incorrect relation.]
Here, the output mentions the relevant entities or values but connects them
using a different predicate. For example, \texttt{Gemma4-31B} received:

\begin{quote}\small
\textit{Triple:}
Ľubomír Lisý $\mid$ hasField $\mid$ neurológia

\textit{Output:}
Ľubomír Lisý je neurológ.\\
\textit{[Ľubomír Lisý is a neurologist.]}
\end{quote}

Having neurology as one's field does not necessarily state that the person is
a neurologist. The output changes a field-of-work relation into an occupation
or identity relation.

\item[\textsc{Ent}: incorrect entity or value.]
This category is used when an entity or object from the source is replaced,
misidentified, or assigned an incorrect value. For example,
\texttt{Tiny-Aya} received:

\begin{quote}\small
\textit{Triple:}
Slovensko $\mid$ numberOfMinistries $\mid$ 15

\textit{Output:}
Slovenia has 15 ministries.
\end{quote}

The number and relation are retained, but \textit{Slovensko} (Slovakia) is
replaced with Slovenia, producing a claim about a different country.

\end{description}

These categories describe the main form of disagreement with the source
triples. They should not be interpreted as fully independent phenomena. For
example, replacing an entity may also make the resulting claim unsupported,
and omitting one argument can lead the remaining relation to be interpreted
incorrectly.

\end{document}